\documentclass[runningheads]{llncs}
\usepackage[T1]{fontenc}
\usepackage{graphicx,verbatim}
\usepackage{subcaption}
\usepackage{booktabs}
\usepackage[dvipsnames]{xcolor}
\usepackage{xspace}
\usepackage{wrapfig}
\usepackage{caption}

\usepackage{multirow}
\usepackage{multicol}
\usepackage{colortbl}
\usepackage{tabularx}
\usepackage{tcolorbox}
\usepackage{etoolbox}
\usepackage{booktabs}
\usepackage{rotating}
\usepackage{soul}
\usepackage{cancel}
\usepackage{amssymb}
\usepackage{amsmath}

\newcommand{\algname}{SimCroP\xspace}

\makeatletter
\DeclareRobustCommand\onedot{\futurelet\@let@token\@onedot}
\def\@onedot{\ifx\@let@token.\else.\null\fi\xspace}

\makeatother

\makeatletter
\DeclareRobustCommand\onedot{\futurelet\@let@token\@onedot}
\def\@onedot{\ifx\@let@token.\else.\null\fi\xspace}

\makeatother

\definecolor{darkgreen}{rgb}{0,0.7,0}
\definecolor{darkyellow}{rgb}{0.8,0.8,0}
\definecolor{bittersweet}{rgb}{1.0, 0.44, 0.37}
\definecolor{amber}{rgb}{1.0, 0.49, 0.0}
\definecolor{lgray}{rgb}{0.83,0.83,0.83}

\definecolor{color_unlabled}{rgb}{0.0,0.0,0.0}
\definecolor{color_vehicle}{rgb}{0.0,0.0,0.56}
\definecolor{color_road}{rgb}{0.5,0.25,0.5}
\definecolor{color_redlight}{rgb}{1.0,0.0,0.0}
\definecolor{color_person}{rgb}{0.859,0.078,0.234}
\definecolor{color_roadline}{rgb}{0.613,0.914,0.195}
\definecolor{color_sidewalk}{rgb}{0.953,0.137,0.906}

\definecolor{ellisred}{rgb}{0.87,0.44,0.38} 
\definecolor{ellisgreen}{rgb}{0.69,0.90,0.52} 
\definecolor{elliscyan}{rgb}{0.29,0.77,0.74} 
\definecolor{ellisorange}{rgb}{0.89,0.55,0.28} 
\definecolor{ellisblue}{rgb}{0.41,0.61,0.86} 

\usepackage{hyperref}
\usepackage{marvosym}
\usepackage{pifont}
\usepackage{color}

\begin{document}
\title{SimCroP: Radiograph Representation Learning with Similarity-driven Cross-granularity Pre-training}
\titlerunning{Similarity-driven Cross-granularity Pre-training}

%
\author{Rongsheng Wang\inst{1,2,6}$^{\star}$ \and
Fenghe Tang\inst{1,2}\thanks{Equal contribution} \and
Qingsong Yao\inst{3} \and
Rui Yan\inst{1,2} $^{\href{mailto:yanrui@ustc.edu.cn}{\textrm{\Letter}}}$ \and \\
Xu Zhang\inst{1,2,6} \and
Zhen Huang\inst{2} \and
Haoran Lai\inst{1,2,6} \and
Zhiyang He\inst{6} \and
Xiaodong Tao\inst{6} \and
Zihang Jiang\inst{1,2} $^{\href{mailto:jzh0103@ustc.edu.cn}{\textrm{\Letter}}}$ \and
Shaohua Kevin Zhou\inst{1,2,4,5} $^{\href{mailto:skevinzhou@ustc.edu.cn}{\textrm{\Letter}}}$} 
\authorrunning{R. Wang, F. Tang et al.}
%
\institute{School of Biomedical Engineering, Division of Life Sciences and Medicine, University of Science and Technology of China (USTC), Hefei Anhui, 230026, China \and
Center for Medical Imaging, Robotics, Analytic Computing \& Learning (MIRACLE), Suzhou Institute for Advance Research, USTC, 215123, China \and
Stanford University, Palo Alto, California, 94025, United States \and
Jiangsu Provincial Key Laboratory of Multimodal Digital Twin Technology, Suzhou \and
Key Laboratory of Precision and Intelligent Chemistry, USTC \and
Anhui IFLYTEK CO., Ltd. \\
}



\maketitle              

%
\begin{abstract}
Medical vision-language pre-training shows great potential in learning representative features from massive paired radiographs and reports. However, in computed tomography (CT) scans, the distribution of lesions which contain intricate structures is characterized by spatial sparsity. 
Besides, the complex and implicit relationships between different pathological descriptions in each sentence of the report and their corresponding sub-regions in radiographs pose additional challenges. 
In this paper, we propose a \textbf{Sim}ilarity-Driven \textbf{Cro}ss-Granularity \textbf{P}re-training (\algname) framework on chest CTs, which combines similarity-driven alignment and cross-granularity fusion to improve radiograph interpretation. 
We first leverage multi-modal masked modeling to optimize the encoder for understanding precise low-level semantics from radiographs. Then, similarity-driven alignment is designed to pre-train the encoder to adaptively select and align the correct patches corresponding to each sentence in reports.
The cross-granularity fusion module integrates multi-modal information across instance level and word-patch level, which helps the model better capture key pathology structures in sparse radiographs, resulting in improved performance for multi-scale downstream tasks. 
\algname is pre-trained on a large-scale paired CT-reports dataset and validated on image classification and segmentation tasks across five public datasets.
Experimental results demonstrate that \algname outperforms both cutting-edge medical self-supervised learning methods and medical vision-language pre-training methods. Codes and models are available at \href{https://github.com/ToniChopp/SimCroP}{https://github.com/ToniChopp/SimCroP}.


\end{abstract}
\section{Introduction}

Deep learning (DL) has demonstrated exceptional potential in the realm of radiograph representation learning~\cite{zhou2021review,pathak2022deep,chen2022review,huang2024pele,shao2023diffuseexpand,huang2025casemark,zhou2024hybrid,hyspark,mambamim,hiend}, which is trained on large-scale annotated datasets and 
achieves performance on par with that of clinical expert. 
However, the annotation of radiographs remains a resource-intensive and burdensome endeavor for clinical practitioners outside their regular duties, posing a significant bottleneck in the advancement of DL applications in medical imaging. Medical vision-language pre-training (Med-VLP)~\cite{huang2021gloria,wang2022mgca} seeks to capitalize on the detailed textual interpretations provided by radiograph-report pairs to assist radiograph representation learning, which has emerged as a prominent focus in contemporary researcher~\cite{tiu2022chexzero,li2024mlip,Wang2023ECAMP,wang-etal-2022-medclip,zhou2023mrm,huang2024maco}.

However, radiographs usually have complex textures and structures, especially for 3D chest CT scans. This primary challenge lies in extracting precise representations from deeper feature spaces~\cite{hatamizadeh2022unetr}. Prior Med-VLP studies on 3D chest CTs have predominantly relied on contrastive learning (CL)~\cite{clip}. For instance, CT-CLIP~\cite{hamamci2024ctclip} introduces a CT-centric contrastive language-image pre-training framework, which optimizes the mutual information between global representations. Similarly, M3D~\cite{bai2024m3d} develops a multi-modal large language model built on CL principles. BIUD~\cite{cao2024biud} bootstraps the understanding of 3D chest CT images by distilling chest-related diagnostic knowledge from an extensively pre-trained 2D X-ray expert model. Moreover, fVLM~\cite{fvlm_iclr25} adopts a fine-grained approach, aligning anatomical regions of CT images with their comparable descriptions in radiology reports and performing CL on each anatomical region individually. MG-3D~\cite{ni2024mg3d} incorporates both intra-patient cross-modal semantic consistency and inter-patient semantic correlations into cross-modal attention mechanisms.

Nevertheless, prevailing methods exhibit notable limitations in effectively utilizing prior knowledge from reports for radiograph representation learning in two critical aspects. First, the spatial distribution of lesions posing intricate structures is characterized by spatial sparsity~\cite{fvlm_iclr25}, presenting considerable difficulties in extracting visual features. 
Second, radiology reports exhibit hierarchical linguistic structures, consisted of descriptive sentences describing correlative visual features and interpretive statements synthesizing the clinical narratives~\cite{hamamci2024ctclip}, introduces inherent complexity that hinders effective pre-training. 
Furthermore, the absence of explicit spatial grounding annotations for descriptive sentences introduces substantial optimization challenges in establishing precise correspondences between sentences and massive visual feature space during pre-training.


To address these issues, we propose {\bf Sim}ilarity-driven {\bf Cro}ss-granularity {\bf P}re-training ({\bf \algname}), which pre-trains strong representation by aligning descriptive sentences in radiology reports and their corresponding sub-regions in radiographs. 
\algname addresses three core medical self-supervised learning (Med-SSL) objectives: 
(1) {\it Masked image modeling}, which enhances the model’s ability to capture structural and textural details in sparse radiographs.
(2) {\it Sentence-subregion alignment}. Inspired by the fact that doctors usually write sentence to describe the pathological context of some subregions in the radiograph, we first choose the sentence level as the granularity for extracting the supervision in the report. Then, we design Similiarity-driven Alignment (SA) to pull the text feature of each sentence closer to the vision features of its most similar visual patches. Without any manual annotations, SA automatically optimizes the vision encoder to select and align the correct patches which are reflect to each sentence in reports.
(3) {\it Cross-granularity masked report modeling}, which integrates instance-level visual features via global average pooling (GAP) with word-patch level cross-modal features, to facilitate the reconstruction of the masked reports. 

To comprehensively evaluate the effectiveness of \algname, we pre-train our Med-VLP framework on the large-scale medical dataset CT-RATE~\cite{hamamci2024ctclip}, which comprises paired chest CT images and reports. We conduct extensive experiments on multi-granularity downstream tasks, including linear probing classification and fine-tuning segmentation, across five public datasets. 
Empirical results demonstrate the superiority and generalization ability of \algname, significantly surpassing state-of-the-art (SOTA) Med-VLP methods with substantial performance improvements.

\section{Method}
\begin{figure}[h!]
    \centering
    \includegraphics[width=\linewidth]{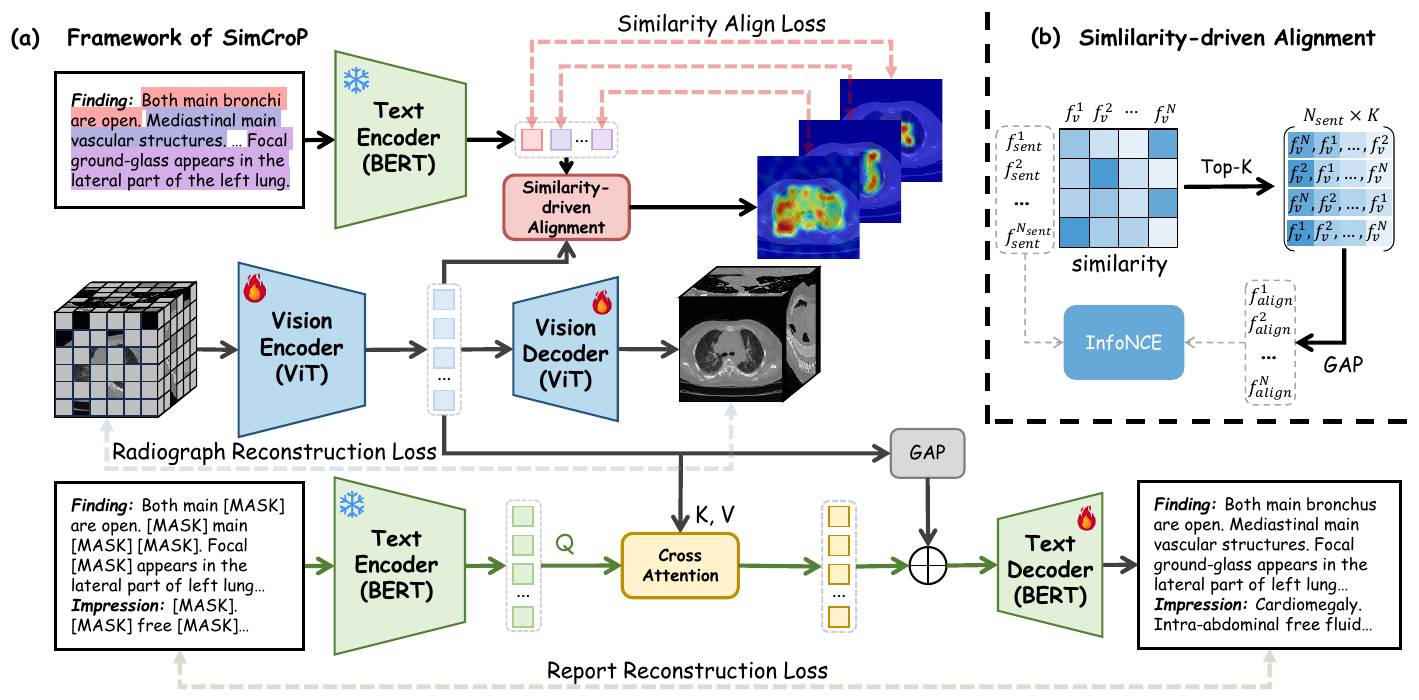}
    \caption{(a) Overall framework of proposed \algname for medical vision-language pre-training. We combine multi-modal masked modeling, similarity-driven alignment, and cross-granularity fusion to achieve effective radiograph representation learning.  (b) Details of similarity-driven alignment. 
    For each descriptive sentence, we calculate the similarity with each patch in the radiograph and select the top-K most similar patches to ensure better alignment. }
    \label{fig:framework}
\end{figure}

In this section, we delve into the design of \algname for medical vision language pre-training on chest CTs. 
Fig.~\ref{fig:framework} illustrates our similarity-driven cross-granularity pre-training (\algname) framework. 
The fire and ice icons represent the parameters of the module that are trained and frozen, with shared weights between the text encoders. First, we briefly introduce the multi-modal masked modeling for 3D radiographs and paired reports utilized in our framework in section~\ref{sec:mask}. Then, we illustrate how \algname leverages similarity-driven alignment for better fit the sparsity of radiographs like CTs in section~\ref{sec:similar}. Finally, we clarify the detailed approaches of cross-granularity fusion in section~\ref{sec:fusion}.

\subsection{Masked Modeling}
\label{sec:mask}
Our approach is grounded on the multi-modal masked autoencoder architecture~\cite{chen2022m3ae}. Following previous works~\cite{bai2024m3d,cao2024biud,fvlm_iclr25}, we adopt vision transformer (ViT)~\cite{vit} and BERT~\cite{lu2019vilbert} as the vision and text encoders, respectively. 

\noindent\textbf{Radiograph masking.} Given a radiograph $I \in \mathbb{R}^{H\times W\times D}$ and its paired report $T$, we first split the radiograph into $\frac{H}{P_H}\times \frac{W}{P_W}\times \frac{D}{P_D}$ patches with size $P_H\times P_W\times P_D$. Following MAE~\cite{He2021MAE}, we mask 75\% of the patches, resulting in $N$ unmasked patches $I_u=\{I_u^s\}_{s=1}^{N}$. These unmasked patches are then augmented with 3D position embeddings $E_{pos}$ as described in ~\cite{zhou2023self} and passed through the vision encoder to obtain the vision feature $f_v=E_v(I_u, E_{pos})$. During the vision decoding stage, $f_v$ is fed into the vision decoder $D_v$ along with mask tokens $f_m$, to reconstruct the radiograph $\hat{I}=D_v(f_v, f_m)$. We optimize the reconstructed radiograph using the mean squared error (MSE) loss:
\begin{equation}
    \mathcal{L}_{\text{MIM}}(\hat{I}, I) = \text{MSE}(D_v(E_v(I_u, E_{pos}), f_m), I).
\end{equation}

\noindent\textbf{Report masking.} For the original report consisting of $N_t$ words, we first tokenize the words to tokens. Then, we randomly mask a proportion ($\gamma$) of tokens from the report $T$, resulting in a masked set of tokens $T_m=\{T_m^l\}_{l=1}^{N_m}$ and the unmasked tokens $T_u=\{T_u^l\}_{l=1}^{N_t-N_m} $. Both masked and unmasked tokens are passed together through the text encoder to obtain the masked report feature $f_{t}=E_t(T_m, T_u)$.  The decoding procedure will be introduced in secion~\ref{sec:fusion}.

\begin{table*}[t]
\small
\centering
\setlength{\tabcolsep}{ 3.3pt}
\caption{ Evaluating our method against other SOTA Med-SSL and Med-VLP approaches on the linear probing classification task with \textbf{3D ViT-B backbone}. 
The best and second-best results are in \textbf{bold} and \underline{underlined}, respectively. }
\begin{tabular}{l|ccc|ccc|cc|cc}
\toprule
\multirow{3}{*}{Method}        & \multicolumn{3}{c|}{CT-RATE}         & \multicolumn{3}{c|}{CC-CCII}  & \multicolumn{2}{c|}{Rad-ChestCT} & \multicolumn{2}{c}{LUNA16}\\ 
& \multicolumn{3}{c|}{(AUC)}  & \multicolumn{3}{c|}{(ACC)}  & \multicolumn{2}{c|}{(AUC)}  & \multicolumn{2}{c}{(AUC)} \\
  & \multicolumn{1}{c}{1\%} & \multicolumn{1}{c}{10\%} & \multicolumn{1}{c|}{100\%} & \multicolumn{1}{c}{1\%} & \multicolumn{1}{c}{10\%} & \multicolumn{1}{c|}{100\%} & \multicolumn{1}{c}{10\%} & \multicolumn{1}{c|}{100\%} & \multicolumn{1}{c}{10\%} & \multicolumn{1}{c}{100\%}\\ \midrule                  
Random init                      &58.2  &60.9  &63.8  &57.8  &74.7  &77.2  &56.4  &60.0  &57.6  &60.4  \\\midrule

\multicolumn{4}{l}{\textit{\textcolor{gray}{3D Med-SSL}}} \\
MAE~\cite{He2021MAE}             &75.6  &78.1  &79.9  &67.4  &81.6  &88.5  &68.6  &71.8  &66.3  &70.5  \\\midrule
\multicolumn{4}{l}{\textit{\textcolor{gray}{3D Med-VLP}}} & \\
M3AE~\cite{chen2022m3ae}         &77.2  &79.5  &81.0  &69.8  &81.7  &89.9  &69.3  &72.2  &\textbf{67.1}  &70.8\\
CT-CLIP~\cite{hamamci2024ctclip} &74.1  &78.6  &80.4  &65.0  &76.7  &83.6  &67.5  &69.1  &64.0  &65.0\\
MRM~\cite{zhou2023mrm}           &77.7  &81.7  &82.1  &70.2  &81.3  &90.3  &\underline{72.1}  &72.6  &66.6  &\underline{72.2}\\
M3D~\cite{bai2024m3d}            &74.3  &79.1  &80.7  &65.4  &77.0  &83.8  &68.0  &69.8  &65.2  &68.4\\
fVLM~\cite{fvlm_iclr25}          &\underline{79.4}  & \underline{81.8}  &\underline{82.2}  &\underline{72.3}  &\underline{82.9}  &\underline{90.7}  &71.1  &\underline{74.2}  &65.9  &71.3\\
\rowcolor[gray]{0.9} \algname (Ours)  &{\bf 81.0}  & {\bf 82.4}  & {\bf 82.9}  & {\bf 73.1}  & {\bf 83.2}  & {\bf 91.3}  & {\bf 73.4}  & {\bf 75.8}  & \underline{67.0}  & {\bf 73.3}\\\bottomrule
\end{tabular}
\label{Table:LinearProbe}
\end{table*}

\subsection{Simlilarity-driven Alignment}
\label{sec:similar}
Given an input report $T=[T_\text{F}, T_\text{I}]$, $T_\text{F}$ and $T_\text{I}$ indicates the \textit{finding} and \textit{impression} section, respectively. We strategically disaggregate the "\textit{Finding}" section $T_\text{F}$ due to its clinical relevance in encapsulating comprehensive visual observations. For a "\textit{Finding}" component comprising $N_{sent}$ linguistically independent clauses, we directly feed the tokenized sentence ensemble $T_{sent}=\{T_{sent}^l\}_{l=1}^{N_{sent}}$ into the text encoder to derive sentence-level embeddings $f_{sent}^l=E_t(T_{sent}^l)$. As illustrated in Fig.~\ref{fig:framework}(b), similarity-driven alignment between the $l$-th sentence $T_{sent}^l$ and  $s$-th unmasked patch $I_u^s$ can be calculated as:
\begin{equation}
    Sim_{l, s} (T_{sent}^l, I_u^s)= [E_t(T_{sent}^l)]^T[E_v(I_u^s, E_{pos})].
\end{equation}
The top-K most relevant spatial correspondences per sentence are then identified through:
\begin{equation}
    Sim_{l, K} (T_{sent}^l, I_u^s)= \mathop{\text{TopK}}\limits_{0\le s<N}Sim_{l, s} (T_{sent}^l, I_u^s).
\end{equation}
These discriminative visual features $f_v^K$ undergo spatial aggregation via GAP to produce similarity-driven aligned feature for the $l$-th sentence $f_{align}^l=\text{GAP}(f_v^K)$. To enforce semantic coherence between subregion visual features and sentence features, we employ similarity-driven fine-grained contrastive learning with the following symmetric InfoNCE loss~\cite{Oord2018RepresentationLW}:
\begin{equation}
    \mathcal{L}_{align}= -\frac{1}{N_{sent}} \sum_{i=1}^{N_{sent}} [\log\frac{\exp(s_{i,i}^{vt}/\tau)}{\sum_{j=1}^{N_{sent}}\exp(s_{i,j}^{vt}/\tau)} + \log\frac{\exp(s_{i,i}^{tv}/\tau)}{\sum_{j=1}^{N_{sent}}\exp(s_{i,j}^{tv}/\tau)}],
\end{equation}
where $s_{i,j}^{vt}=(f_{align}^i)^Tf_{sent}^j, s_{i,j}^{tv}=(f_{sent}^i)^Tf_{align}^j$, $\tau$ denotes the temperature, which is set to 0.07 following common practice.

\subsection{Cross-granularity Fusion}
\label{sec:fusion}
Diverging from M3AE~\cite{chen2022m3ae}, which singularly employs cross attention mechanisms to aggregat vision feature $f_v$ to the report feature $f_t$, we propose a hierarchical fusion architecture comprising two complementary components: 
\begin{itemize}
    \item Instance-level vision features $f^I$ are derived through global average pooling: $f^I=GAP(f_v)$;
    \item Word-patch level cross-modal features $f^{W}$ are computed via scaled dot-product cross attention:
\end{itemize}
\begin{equation}
    f^{W} = \text{CrossAttention}(f_t, f_v)=\text{Softmax}(\frac{f_tW_q(f_vW_k)^T}{\sqrt{d_k}})f_vW_v,
\end{equation}
where $W_q, W_k$, and $ W_v$ are linear projection matrix, $d_k$ is the feature dimension. Finally, the fused features are passed through the text decoder to reconstruct the masked tokens $\hat{T}_m=D_t(f^I + f^{W})$. The masked text modeling loss for report reconstructing can be formulated as 
\begin{equation}
    \mathcal{L}_{\text{MLM}} = - \frac{1}{N_m} \sum_{l = 1}^{N_m} \log P(\hat{T}_m^l = T_m^l \mid \hat{T}_m).
\end{equation}
The overall loss function of \algname is as follows:
\begin{equation}
    \mathcal{L}= \mathcal{L}_{\text{MIM}} + \lambda_1\mathcal{L}_{align} + \lambda_2\mathcal{L}_{\text{MLM}}.
\end{equation}
Empirically, we configure $\lambda_1=\lambda_2=1$ to maintain equilibrium between alignment and reconstruction objectives.

\begin{table}[t]
\begin{minipage}[t]{0.52\textwidth}
    \centering
    \setlength{\tabcolsep}{ 3pt}
\caption{ Segmentation results of Med-SSL and Med-VLP approaches on the fine-tuning segmentation task with \textbf{3D ViT-B backbone}. 
The best and second-best results are in \textbf{bold} and \underline{underlined}, respectively. }
\begin{tabular}{l|cc|c}
\toprule
\multirow{3}{*}{Method}        & \multicolumn{2}{c|}{LUNA16}         & \multicolumn{1}{c}{BTCV}  \\ 
&\multicolumn{2}{c|}{(Dice)} & \multicolumn{1}{c}{(Dice)}\\
& 10\% &100\% & 100\%\\\midrule
Random init                     &91.9  &92.3  &78.9  \\\midrule
\multicolumn{3}{l}{\textit{\textcolor{gray}{3D Med-SSL}}} \\
MAE~\cite{He2021MAE}             &92.7  &93.3  &80.3  \\\midrule
\multicolumn{3}{l}{\textit{\textcolor{gray}{3D Med-VLP}}} \\
M3AE~\cite{chen2022m3ae}    &93.1  &\textbf{93.7}  &\underline{80.5}\\
CT-CLIP~\cite{hamamci2024ctclip}  &92.9  &93.4  &79.4\\
MRM~\cite{zhou2023mrm}           &\underline{93.2}  &\underline{93.5}  &80.3\\
M3D~\cite{bai2024m3d}       &92.9  &93.2  &79.6\\
fVLM~\cite{fvlm_iclr25}    &93.1  &93.4  &80.0\\
\rowcolor[gray]{0.9} \algname (Ours)      &\textbf{93.5}  &\textbf{93.7}  & \textbf{80.7}\\\bottomrule
\end{tabular}
\label{Table:Segmentation}
  \end{minipage}
  \hfill
  \begin{minipage}[t]{0.45\textwidth}
\setlength{\tabcolsep}{ 2pt}
\caption{ Ablation study on each design component in our framework on linear-probe classification and fine-tuning segmentation. "SA" refers to Similarity-driven Alignment, "IL" denotes Instance-level vision features, and "WL" stands for Word-patch level cross-modal features. $\checkmark$ and $\times$ denote whether the component is included.}
\label{Table:ablation}
\begin{tabular}{ccc|c|c}
\toprule
\multirow{3}{*}{SA}  &\multirow{3}{*}{IL}  &\multirow{3}{*}{WL}  &\multicolumn{1}{c|}{RadChestCT}         &\multicolumn{1}{c}{LUNA16}  \\ 
&& &\multicolumn{1}{c|}{(AUC)} &\multicolumn{1}{c}{(Dice)} \\
&& & 10\% & 10\%\\\midrule
$\times$ &$\checkmark$  &$\times$           &71.6  &92.0  \\
$\times$ &$\times$  &$\checkmark$           &69.8  &91.7  \\
$\times$ &$\checkmark$  &$\checkmark$       &72.4  &92.4  \\\midrule
$\checkmark$  &$\checkmark$  &$\times$      &73.3   &92.8  \\
$\checkmark$  &$\times$  &$\checkmark$      &72.7   &92.6  \\
$\checkmark$  &$\checkmark$  &$\checkmark$  &\textbf{73.4}   &\textbf{93.5}  \\\bottomrule
\end{tabular}



\end{minipage}
\end{table}

\section{Experiments and Results}
\noindent\textbf{Pre-training datasets.}
We employ CT-RATE~\cite{hamamci2024ctclip}, a public large-scale dataset comprising 50,188 CT volumes with paired reports. Per official split, we leverage 47,149 volume-report pairs from the official training subset during pre-training.

\noindent\textbf{Fine-tuning datasets.}
CT-RATE~\cite{hamamci2024ctclip} and RadChestCT~\cite{draelos2021radchestct} establish multi-label classification benchmarks with official data partition. 
The CC-CCII~\cite{he2020cc-ccii} dataset addresses multi-class pneumonia classification task with a 7:3 random partition. LUNA16~\cite{setio2017luna16} provides dual objectives of nodule classification and pulmonary segmentation, with analogous 7:3 randomized data stratification. To assess cross-domain transferability of learned chest CT representations to abdominal imaging, we employ the BTCV~\cite{landman2015btcv} dataset under its prescribed organ segmentation protocol with official data splits. Unless official validation set is defined, testing sets serve as validation sets for all experimental configurations.

\noindent\textbf{Implementation.} 
We resample the volume to $1.5\times1.5\times3.0$ spacing using tri-linear interpolation and map the Hounsfield unit range from $(-1000, 1000)$ to $(-1, 1)$, with clipping. The image size is set to be $224\times224\times112$. We leverage an 8-layer 3D ViT-B~\cite{vit} as vision encoder initialized with MAE ImageNet-1K pre-trained weights~\cite{He2021MAE} and a 4-layer 3D ViT-B for vision decoder, with patch size of $16\times16\times8$. Additionally, we employ pre-trained CXR-BERT~\cite{boecking2022cxrbert} as our text encoder and a 6-layer BERT as our text decoder. The model is trained for 140 epochs on A800 GPUs with a batch size of 48, using AdamW as the optimizer with a learning rate of 1.5e-4 and weight decay of 0.05. 

\noindent\textbf{Baselines.}
We compare \algname with one Med-SSL method, MAE~\cite{He2021MAE}, and five Med-VLP methods: M3AE~\cite{chen2022m3ae}, CT-CLIP~\cite{hamamci2024ctclip}, MRM~\cite{zhou2023mrm}, M3D~\cite{bai2024m3d}, and fVLM~\cite{fvlm_iclr25}. For fair comparison, we directly use the official pre-trained weights of M3D and re-implement the other methods under the same data settings as our approach.

\begin{figure}[t]
\begin{minipage}[t]{0.55\textwidth}
    \centering
    \includegraphics[width=\linewidth]{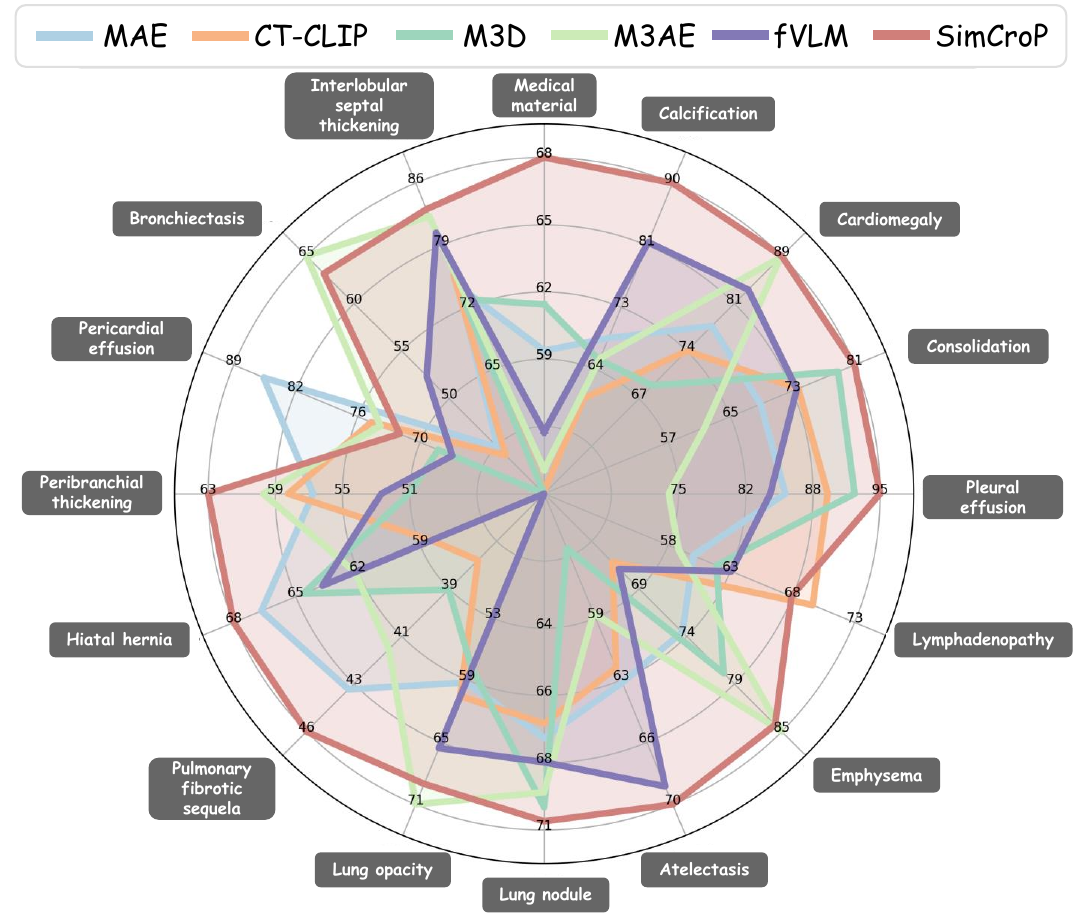}
    \caption{Test results for each label on external dataset RadChestCT.}
    \label{fig:external_cls}
\end{minipage}
\hfill
\begin{minipage}[t]{0.4\textwidth}
    \centering
    \includegraphics[width=\linewidth]{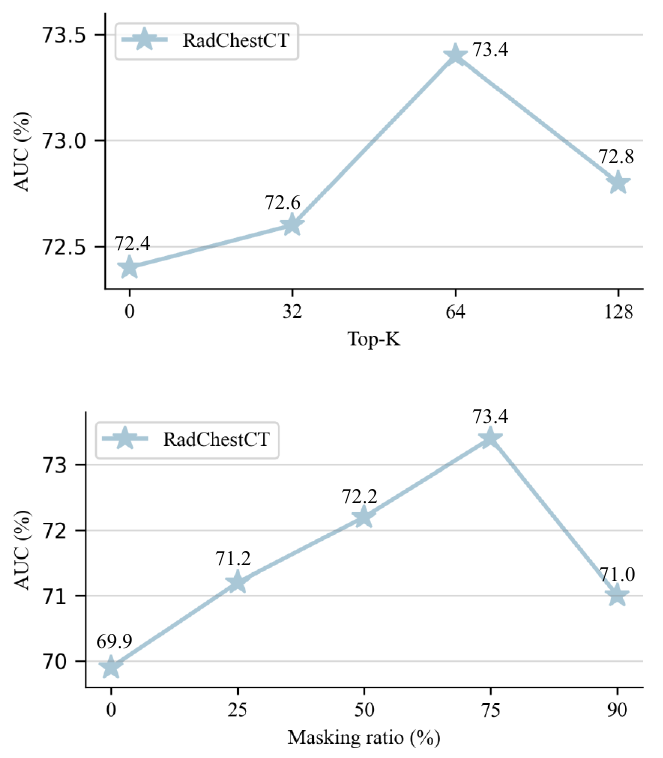}
    \caption{Ablation on Top-K and report masking ratio selection.}
    \label{fig:abla_cls}
\end{minipage}
\end{figure}

\noindent\textbf{Classification results.}
We perform linear-probe classification with our \algname on four datasets, 
as shown in Table~\ref{Table:LinearProbe}, our \algname significantly outperforms 3D Med-SSL and 3D Med-VLP approaches across different training data ratios on all four datasets. 
Notably, on CT-RATE, which includes the largest number of disease labels and test data, \algname surpasses the current state-of-the-art (SOTA) method by 1.6\% AUC while using 1\% labeled data. 
Interestingly, we find the masked modeling-based methods M3AE~\cite{chen2022m3ae}, MRM~\cite{zhou2023mrm} and \algname outperform contrastive-based methods like M3D~\cite{bai2024m3d} and fVLM~\cite{fvlm_iclr25} especially on fine-grained nodule classification task introduced by LUNA16. Additionally, Fig.\ref{fig:external_cls} shows the test results for each label on the external dataset Rad-ChestCT, while training on 100\% of the training data from CT-RATE. The label transfer setting follows CT-CLIP\cite{hamamci2024ctclip}, and the transfer results further demonstrate the robustness and superiority of the representation learned by \algname.

\noindent\textbf{Segmentation results.}
To evaluate the effectiveness of fine-grained radiograph representations learned by \algname, fine-tuning segmentation tasks are conducted in Table~\ref{Table:Segmentation} using UNETR~\cite{hatamizadeh2022unetr} framework. \algname consistently outperforms all aforementioned methods across lung segmentation and abdominal organ segmentation. This demonstrates that \algname effectively leverages its strong ability to utilize cross-granularity information between radiographs and reports. Therefore, \algname generalizes well across different medical vision tasks, establishing itself as a robust and highly effective Med-VLP method.

\noindent\textbf{Ablation study.}
As demonstrated in Table~\ref{Table:ablation}, our similarity-driven alignment module achieves substantial performance gains in radiograph representation learning, with improvement differentials exceeding 1\% and 1.1\% on multi-label classification and pulmonary segmentation task respectively. A critical observation reveals that the ablation configuration devoid instance-level visual features underperforms the variant excluding word-patch cross-modal features for lung segmentation. This phenomenon can be attributed to the domain-specific characteristic of chest CTs where macro-anatomical structures occupy significantly larger voxel distributions compared to localized pathologies, thereby rendering holistic instance-level features more discriminative for lung segmentation.
In Fig.~\ref{fig:abla_cls}, ablation studies investigating the selection of top-K and report masking ratios demonstrate that subregions comprising of 64 patches (appropriately 10\% of the radiograph area) optimally align with the semantic granularity of descriptive sentences. This empirical evidence validates \algname's capability to address the inherent spatial sparsity of radiographs. Furthermore, a report masking ratio of 75\% determines to maximize radiograph representation learning efficacy.

\begin{figure}[t]
    \centering
    \includegraphics[width=\linewidth]{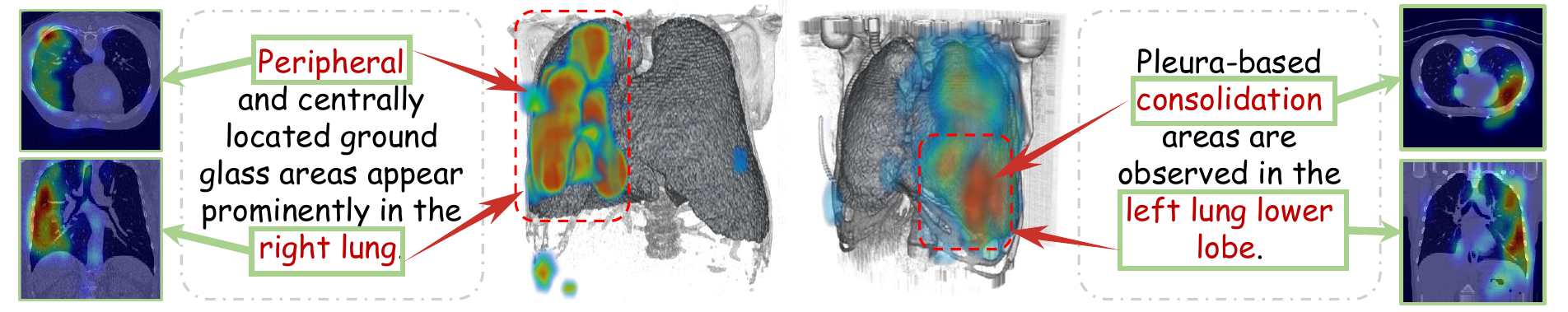}
    \caption{Visualization of top-K patches correlated to descriptive sentences.}
    \label{fig:vis}
\end{figure}

\noindent\textbf{Representation visualization.}
Fig.~\ref{fig:vis} exhibits the subregions consisted of top-K similar patches correlated to the given descriptive sentences. Lesions and anatomy locations with higher similarity to the sentences are highlighted in red, showcasing \algname's superiority of similarity-driven alignment design.

\section{Conclusion}
We propose a novel medical vision-language pre-training method specifically designed for radiographs with inherent sparsity. 
To address the challenge of spatial sparsity and complex fine-grained connections between sentences and subregions, we pre-train the model to select and align each descriptive sentences with corresponding subregions. Furthermore, to tackle the issue of the massive visual feature space in 3D radiographs, we introduce a cross-granularity fusion module that simultaneously aggregates instance-level and word-patch level features. The effectiveness of our proposed modules is validated through multi-scale downstream tasks, including linear-probe classification and fine-tuning segmentation across multiple datasets, demonstrating the robustness of our method. However, the absence of instance-level cross-modal alignment hinders the zero-shot performance of our approach. 
%
%
In future work, we aim to address this limitation, further enhancing \algname as a more powerful foundation model.

    

\begin{credits}
\subsubsection{\ackname} This work was supported by National Natural Science Foundation of China (62271465, 62402473), Suzhou Basic Research Program under Grant SYG202338, 
and Jiangsu Province Science Foundation for Youths (NO. BK20240464).

\subsubsection{\discintname}
The authors have no competing interests to declare that are relevant to the content of this article.
\end{credits}

%
%
%
\bibliographystyle{splncs04}
\bibliography{main}
\end{document}